\documentclass{article}
\pdfoutput=1

\usepackage{arxiv}
\usepackage[utf8]{inputenc} 
\usepackage[T1]{fontenc}    
\usepackage{hyperref}       
\usepackage{url}            
\usepackage{booktabs}       
\usepackage{amsfonts}       
\usepackage{nicefrac}       
\usepackage{microtype}      
\usepackage{lipsum}
\usepackage{graphicx}
\usepackage{notoccite}
\usepackage{subcaption}
\usepackage{array}
\usepackage{multirow} 
\usepackage{graphicx}
\usepackage{adjustbox}
\usepackage{wrapfig}
\usepackage{mathpazo} 
\usepackage[hang,flushmargin]{footmisc}
\usepackage[style=numeric,sorting=none]{biblatex} 
\usepackage{amsmath}
\usepackage{makecell}
\usepackage{units}
\usepackage{appendix}

\hypersetup{hidelinks} 
\title{LLMs left, right, and center: Assessing GPT's capabilities to label political bias from web domains}

\date{October 22, 2024}

\author{ \href{https://orcid.org/0009-0005-0323-8326}{\includegraphics[scale=0.06]{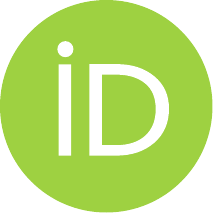}\hspace{1mm}Raphael Hernandes}\\
	Leverhulme Centre for the Future of Intelligence\\University of Cambridge\\
	\texttt{rhh43@cam.ac.uk}\\
    \And 
    \href{https://orcid.org/0000-0002-5130-2258}{\includegraphics[scale=0.06]{orcid.pdf}\hspace{1mm}Giulio Corsi}\\ 
    Leverhulme Centre for the Future of Intelligence\\University of Cambridge\\
    \texttt{gc540@cam.ac.uk} 
}

\renewcommand{\headeright}{}
\renewcommand{\undertitle}{}
\renewcommand{\shorttitle}{LLMs left, right, and center}

\hypersetup{
pdftitle={LLMs left, right, and center: Assessing GPT's capabilities to label political bias from web domains},
pdfsubject={ethics of ai},
pdfauthor={Raphael Hernandes},
pdfkeywords={gpt-4, large language models (LLM), media bias, data labeling, political bias},
}

\begin{document}
\maketitle
\begin{abstract}
This research investigates whether OpenAI's GPT-4, a state-of-the-art large language model, can accurately classify the political bias of news sources based solely on their URLs. Given the subjective nature of political labels, third-party bias ratings like those from Ad Fontes Media, AllSides, and Media Bias/Fact Check (MBFC) are often used in research to analyze news source diversity. This study aims to determine if GPT-4 can replicate these human ratings on a seven-degree scale ("far-left" to "far-right"). The analysis compares GPT-4's classifications against MBFC's, and controls for website popularity using Open PageRank scores. Findings reveal a high correlation ($\text{Spearman's } \rho = .89$, $n = 5,877$, $p < 0.001$) between GPT-4's and MBFC's ratings, indicating the model's potential reliability. However, GPT-4 abstained from classifying approximately $\frac{2}{3}$ of the dataset. It is more likely to abstain from rating unpopular websites, which also suffer from less accurate assessments. The LLM tends to avoid classifying sources that MBFC considers to be centrist, resulting in more polarized outputs. Finally, this analysis shows a slight leftward skew in GPT's classifications compared to MBFC's. Therefore, while this paper suggests that while GPT-4 can be a scalable, cost-effective tool for political bias classification of news websites, its use should be as a complement to human judgment to mitigate biases.
\end{abstract}


Is \textit{The New York Times} more politically inclined to the left or the right? The answer might vary according to the respondent. However, it seems to converge to the center-left ---at least according to three different services that specialize in rating the news: Ad Fontes Media, AllSides, and Media Bias/Fact Check (MBFC)~\cite{AllSidesMediaBias2024}, \cite{InteractiveMediaBias2023}, \cite{NewYorkTimes2024}. Finding that sort of information about one of the most-read news outlets in the world is relatively easy, especially given that its public editor acknowledges the newspaper is perceived as left-wing~\cite{spaydWhyReadersSee2016}. It might be harder, however, to assess the political bias of more obscure or niche sources where accessible and accurate information is lacking.

While political labels carry a certain degree of subjectivity, third-party bias ratings may help balance information ecosystems by allowing a better understanding of what is consumed, which is particularly relevant since news outlets do not generally declare their standings~\cite{sheridanShouldYouTrust2021}. This labeling is often used in academic research. Examples include assessing whether news content highlighted by Google Search has political leanings~\cite[8-9]{trielliSearchNewsCurator2019a} and spotting an increased polarization on Twitter in the discussions about the United Nations Conference of the Parties on Climate Change~\cite{falkenbergGrowingPolarizationClimate2022}.

Nonetheless, making such judgments can be time- and resource-consuming. AllSides, for instance, uses a mix of methods, including editorial and panelist reviews, consumer surveys, research from other sources, and community feedback~\cite{HowAllSidesRates2016}. Things get trickier when analyzing in bulk, labeling hundreds of sources.

Artificial Intelligence (AI) might be an alternative for the labeling process, and recent studies have shown high accuracy when employing Large Language Models (LLMs) for data annotation, including within political contexts~\cite{chiangCanLargeLanguage2023}, \cite{gilardiChatGPTOutperformsCrowd2023}, \cite{tornbergChatGPT4OutperformsExperts2023a}, \cite{wuLargeLanguageModels2023}. While some recent studies use such models to label news content based on web domain (to assess credibility~\cite{yangLargeLanguageModels2023}, for example), none explored judging political bias based on URLs.

Considering this scenario, this research investigates whether OpenAI's GPT-4, the state-of-the-art language model according to relevant benchmarks~\cite{HolisticEvaluationLanguage2024}, \cite{MMLUBenchmarkMultitask2024}, can reliably classify news sources based solely on web domain on a seven-degree scale ("far-left," "left," "center-left," "no bias," "center-right," "right," and "far-right"). It follows the questions:

\begin{itemize}
    \item[RQ1] Can an LLM, using only its embedded knowledge with no internet access, classify the political bias of news sources based solely on their URLs in alignment with humans? 
    \item[RQ2] Does classification accuracy vary depending on the popularity of the news source? 
\end{itemize}

In addressing the first question, this paper analyzes the correlation between political ideology ratings attributed by GPT and MBFC. The second question relates to the functioning of these models, which involves discerning patterns in extensive text corpora~\cite{rooseHowDoesChatGPT2023}. The hypothesis is that how often a news source appears in this data might affect the model's performance. 

\section{Background}

\subsection{GPT-4 and LLMs}

GPT-4 is the latest of OpenAI's family of LLMs~\cite{GPT42023}. It was chosen due to GPT's popularity in the AI community since the launch of ChatGPT in November 2022~\cite{huChatGPTSetsRecord2023}. It is the best option available\footnote{Google's Gemini Ultra had a higher score but was not publicly available at the time of writing.} for the tasks analyzed in this research according to MMLU~\cite{MMLUBenchmarkMultitask2024}, a widely used benchmark for models in zero-shot settings~\cite{hendrycksMeasuringMassiveMultitask2021}, \cite{maslejArtificialIntelligenceIndex2023}, and HELM, a comprehensive benchmark for language tasks~\cite[]{HolisticEvaluationLanguage2024}, \cite[]{liangHolisticEvaluationLanguage2023}, \cite[114]{maslejArtificialIntelligenceIndex2023}.

These sorts of AI are created after analyzing the patterns in immense amounts of textual data. They work by predicting combinations of tokens and reply to questions made by users conversationally, besides following instructions based on their input (called a prompt)~\cite{omiyeLargeLanguageModels2023}, \cite{rooseHowDoesChatGPT2023}. LLMs can perform a myriad of language-processing tasks, such as answering questions, writing, summarizing texts, and translating~\cite[99-108]{maslejArtificialIntelligenceIndex2023}.

As these models grow in terms of computation, number of parameters, and training dataset size, they show capabilities that go beyond their training scope and perform well in tasks that are not their primary goal. These are known as emergent properties. They are not entirely understood but seem related to model size~\cite[2-4]{weiEmergentAbilitiesLarge2022}. The lack of clarity about them is an issue, as little is known about their limitations~\cite[4]{tornbergChatGPT4OutperformsExperts2023a}.

\subsection{LLMs for Annotation Tasks}

Among LLMs' emergent properties are applications related to annotating datasets. LLMs were successfully used to assess the quality of texts written by humans and robots regarding grammar, cohesiveness, likability, and relevance~\cite{chiangCanLargeLanguage2023}. Research also shows that GPT in a zero-shot setting can classify tweets according to several criteria, such as whether they are a content moderation issue or relate to pre-defined political topics, with higher accuracy than non-expert humans under the same instructions~\cite{gilardiChatGPTOutperformsCrowd2023}.

In a similar context to this paper, research found a strong correlation ($\text{Spearman's } \rho = .54$, $p < 0.001$) between human and GPT-assigned ratings for news outlets' credibility in a corpus of over 7,000 web domains~\cite{yangLargeLanguageModels2023}. The result is based on an aggregate score of multiple services that evaluate news domain quality rather than a single classifying service~\cite{linHighLevelCorrespondence2023}. These services correlate with each other on varying degrees ($\text{Spearman's } \rho$ range: \texttt{.32-.90}), so GPT would fall in the middle of that distribution. This motivated the investigation of whether GPT could repeat the same level of performance in a political bias classification scenario.

\subsubsection{LLMs in Political Analysis}

LLMs have also displayed promising results in tasks related specifically to political analysis. GPT could accurately classify the political affiliation of a Twitter user based on the content of a single anonymized tweet~\cite{tornbergChatGPT4OutperformsExperts2023a}. The analysis used posts made by all US senators, providing an unequivocal basis for evaluating the answers. The AI performed better than both expert and non-expert humans.

GPT has also shown remarkable capacity to classify US senators in terms of their liberal-conservative ideology, support of gun control, and support of abortion rights, based only on their names and parties using pairwise comparisons~\cite[2]{wuLargeLanguageModels2023}. The rankings are not simply mimicking other scales, contrasting with the idea that these systems might be parroting patterns from their training data~\cite{benderDangersStochasticParrots2021}. If they were merely copying information from elsewhere, it would be simpler to use the original data. Instead, the investigation shows that the ratings came from a mix of senators' behaviors and how these politicians are perceived~\cite[19]{wuLargeLanguageModels2023}. This indicates that LLMs' capabilities in political classification tasks warrant further investigation, as they might offer not only a cheaper and faster way of labeling the data but also a new scale altogether.

\subsection{Rating News Sources' Political Bias}\label{sec:ratingpoliticalbias}

Perception and behavior are currently used forms of rating political bias in media sources. Scales from the likes of AllSides and polling institutes, such as Pew Research Center, classify news sources based on how consumers perceive them~\cite{HowAllSidesRates2016}, \cite{amymitchellPoliticalPolarizationMedia2014}. Language (behavior) has been used in research to classify newspapers: phrases connected to the right meant right-leaning tendencies~\cite{gentzkowWhatDrivesMedia2010}. The measurement correlated to the political incline of readers, given the preferences in the zip codes where the newspapers circulated~\cite[64]{gentzkowWhatDrivesMedia2010}. A study used the self-reported ideology of Facebook users to classify news sources by leveraging popularity among the right or the left to gauge the outlet's political ideology~\cite{bakshyExposureIdeologicallyDiverse2015a}.

MBFC, the baseline in this paper, primarily analyzes news source quality. Its database includes information beyond credibility, such as the source's country and media type (newspaper or TV, for example). It rates political bias by examining editorial content by their position on general philosophy, abortion, economic policy, education policy, environmental policy, gay rights, gun rights, health care, immigration, military, personal responsibility, regulation, social views, taxes, voter ID laws, and workers' rights~\cite{LeftVsRight2021}. It is cited as a source for asserting news quality both in Academia and in the news~\cite[2]{akerPredictingNewsSource2019}, \cite[]{NoEvidenceDisease2021}, \cite[4]{resnickIffyQuotientPlatform2018}.

\subsubsection{Applications of these Ratings}

Analyzing the political bias of news sources is helpful in gauging diversity in the news ecosystem. By classifying ideology and relating it to zip codes, it is possible to note that political ideology in newspapers is more related to appealing to their audience than ownership diversity, as outlets sharing an owner have diverging political stances depending on their readership. This is relevant for regulating the industry (deciding whether to limit ownership, for example)~\cite{gentzkowWhatDrivesMedia2010}. These can also be used to measure diversity in news exposure, when, for example, news sources' political inclinations to analyze the results offered by Google's algorithm. The classification allowed researchers to identify it was slightly skewed to the left, which might direct readers' attention that way~\cite{trielliSearchNewsCurator2019a}.

\section{Methodology}

The analysis gathered a dataset with classifications from MBFC and popularity ratings from Open PageRank. First, the data for every report made by MBFC was downloaded, removing those that did not include information on political bias, such as the ones focusing on scientific content. The web domains from MBFC's database were cleaned to remove inconsistencies (invalid URLs, for example), and missing values were fetched manually from the reports. Two pairs of records shared the same domain, and only one of each was kept. Three records were related to Facebook pages and were removed.

The dataset includes 5,877 observations extracted from MBFC. The bias column from GPT and MBFC was coded into a scale ranging from -3 to +3, meaning "far-left," "left," "center-left," "center," "center-right," "right," and
"far-right." This makes it possible to calculate both the difference and absolute difference between the two measurements ($\text{GPT Bias} - \text{MBFC Bias}$). See the appendices for summary statistics.

Both MBFC's and GPT-4's Bias Scales display a pattern that deviates from a normal distribution (Figure 1), so a non-parametric correlation method, Spearman's \(R\), was used for the comparison. MBFC has more right than left-wing classifications. This is acknowledged by the authors, who explain that their set was mostly balanced until they started taking user requests for websites to rate, which are generally for right-leaning domains~\cite{FrequentlyAskedQuestions2023}.

The analysis primarily focuses on the correlation between the ratings assigned by GPT and MBFC, an approach similar to previous work that compared the LLM's ability to judge news outlet credibility~\cite{yangLargeLanguageModels2023}. There are multiple ways of classifying news sources' political bias, and the exact placement in a scale might vary (as there is no clear line separating center-left from left, for example). Thus, this work is concerned with whether GPT will place a source more to the right when MBFC does it instead of checking whether it predicts the exact value, so the correlation is a better measurement. The system's accuracy is also analyzed later through linear and logistic regressions controlling for website popularity and political stance.

\subsection{Open PageRank}

Open PageRank is a free initiative that applies metrics like the ones used by Google to rank websites, enabling comparison among domains through a scale from 0 to 10 to the decimal level where higher can be interpreted as a proxy for more popular. It provides data about the top 10 million websites in its database~\cite{WhatOpenPageRank2023}. The root domain of each URL ("www.example.com/abc" $\rightarrow$ "example.com") is used to retrieve ratings from Open PageRank's API. Therefore, this analysis does not differentiate between sources that share a root URL ("theguardian.com/observer" and "theguardian.com," for example), meaning popularity might be slightly exaggerated for some. These, however, are not frequent in the data, which contains 5,684 unique root URLs. The distribution of Open PageRank scores is available in the appendices.

\subsection{Prompting GPT-4}\label{sec:prompting}

The GPT version used is \texttt{gpt-4-1106-preview}, the most recent available when writing, queried through OpenAI's API. It was released on Nov. 06, 2023, with up-to-date information until April 2023, and is described by OpenAI as having "improved instruction following" capabilities~\cite{NewModelsDeveloper2023}. Being in "preview" means it does not support high traffic~\cite{OpenAIPlatform2024}, which is not the case in this work. It was cheaper than older versions of GPT-4 (US\$0.01/1k input and US\$0.03/1k output tokens vs. US\$0.03/1k input and US\$0.06/1k output tokens)~\cite{Pricing2024}. It can reply in \textit{JSON} format instead of text, making parsing easier as it removes the need to extract the output from a paragraph.

\texttt{Gpt-4-1106-preview} has a seed parameter to make the model return consistent completions, which is useful for reproducible outputs~\cite{NewModelsDeveloper2023}. The \textit{seed} was set to \texttt{123}. The \textit{temperature} was \texttt{0.0} to reduce randomness and support reproducibility. A low temperature may be preferable for annotation tasks as it increases consistency without decreasing accuracy~\cite[2]{gilardiChatGPTOutperformsCrowd2023}.

The prompts asked the AI to rate the political bias of a URL or return "-1" if it could not classify it. After the main experiment, to better understand the machine's behavior, additional prompts for a random sample asked GPT to justify the output. Sample prompts and responses are available in the appendices. It was used in a zero-shot setting in an attempt to evoke latent information in the model's training. The request specified that the response be in JSON and provided an example of formatting. A system message added context telling GPT to act as an assistant to determine the political bias of websites.

\section{Results}\label{sec:results}

\subsection{Correlation Between GPT-4 and Human-Assigned Political Bias Ratings}\label{sec:correlation}

\begin{figure}[htbp]
    \centering
    \includegraphics[width=\textwidth]{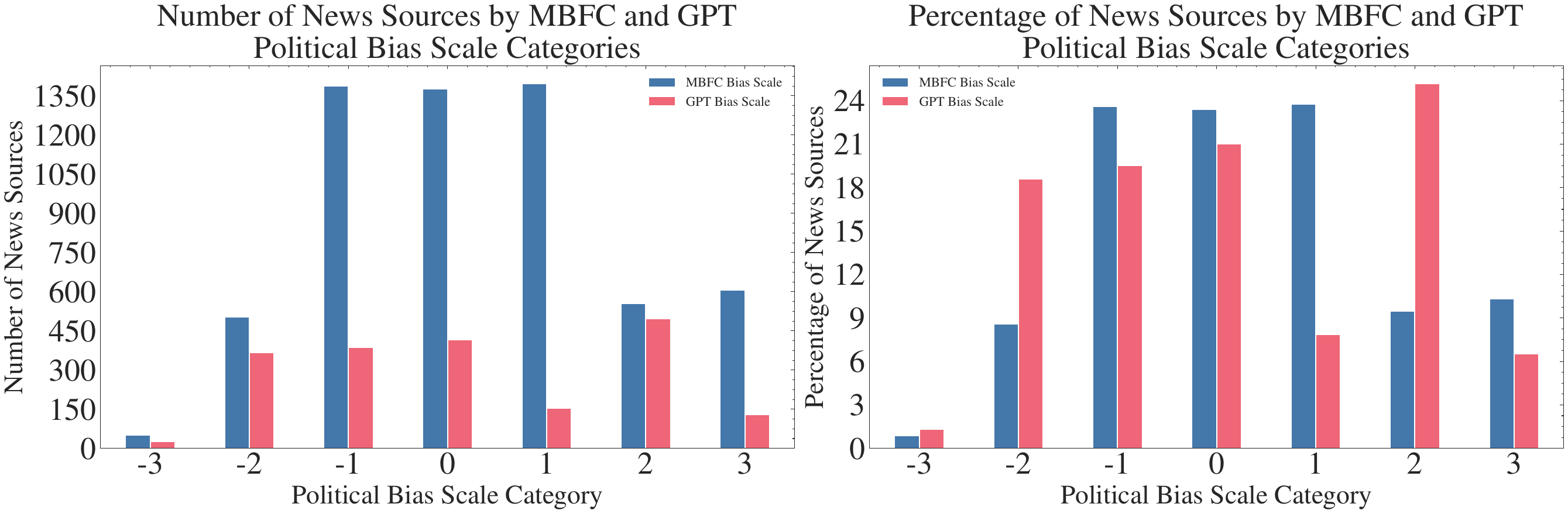}
    \caption{Distribution of news sources in absolute (left) and relative values (right).}
    \label{fig:distribution}
\end{figure}

\begin{figure}[htbp]
    \centering
    \includegraphics[width=\textwidth]{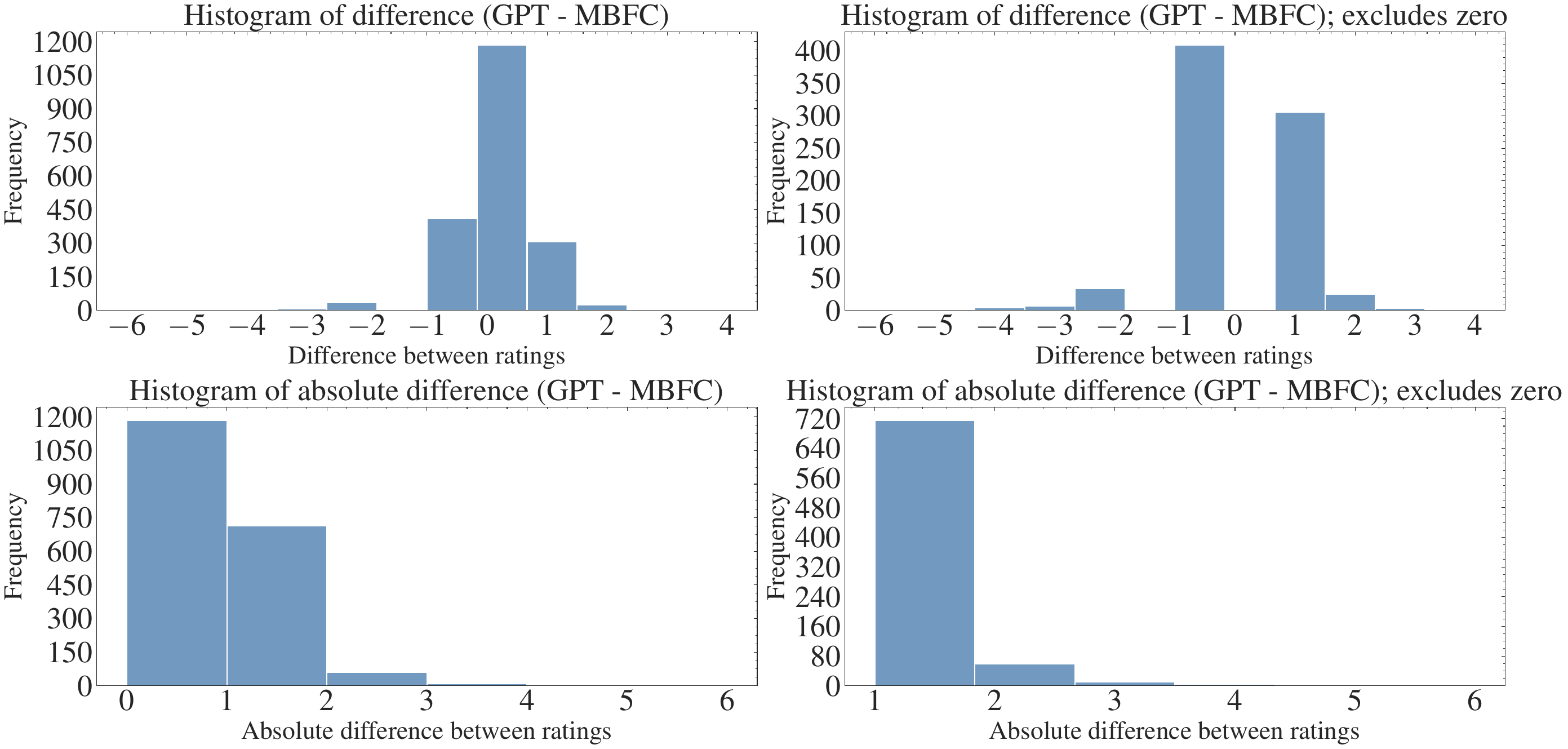}
    \caption{Histograms of difference (top) and absolute difference (bottom) between GPT and MBFC ratings show concentration around minimal difference; charts on the right exclude zero for easier visualization.}
    \label{fig:histograms}
\end{figure}

The analysis shows a very strong\footnote{Based on Quinnipiac University's table for interpreting Spearman's correlation coefficients in Politics contexts~\cite[92]{akogluUserGuideCorrelation2018}} correlation between GPT-4's and MBFC's political bias classifications ($\text{Spearman's } \rho = .89$, $n = 5,877$, $p < 0.001$). This shows that both vary in the same direction: sources classified toward one side of the spectrum in MBFC are likely to follow the same pattern with GPT.

GPT returned a small set of values on the opposite side of the expected spectrum (Figures 3 and 4). The most extreme scenario (far-right being tagged as far-left, or vice-versa) only happened once: \textit{strategic-culture.org} was labeled far-right by MBFC, and GPT had far-left.

The correlation between GPT-4 and MBFC's classifications is at least strong ($> .4$) across most statistically relevant categories. A full breakdown of the categories is available in the appendices. In terms of country, it was stronger in ones that speak English and weakest in unknown regions ($\text{Spearman's } \rho = .49$, $n = 275$) and others ($\rho = .65$, $n = 452$). The correlation was statistically relevant in all locations ($p < 0.001$).

The correlation also fluctuated regarding media type, though staying above the strong threshold for all statistically relevant categories. It was particularly strong among sources not necessarily in the media business: Journals ($\text{Spearman's } \rho = .97$, $n = 7$, $p < 0.01$) and Organization/Foundation ($\text{Spearman's } \rho = .91$, $n = 587$, $p < 0.001$). Most of these are correctly rated "unbiased," which makes sense given the nature of these publications ---less political than a news website. This shows hints of a capacity in GPT-4 to identify these sources as such and classify them accordingly. Across news agencies, the correlation was weak ($\text{Spearman's } \rho = .28$, $n = 41$), though not statistically significant.

Correlation drops when crunching the different political biases into left, right, and center (including center-left and center-right). It is negligible when considering only sources originally classified as left ($\text{Spearman's } \rho = .12$, $n = 556$, $p < 0.05$) and remains strong with both right ($\text{Spearman's } \rho = .43$, $n = 1,162$, $p < 0.001$) and center-leaning sources ($\text{Spearman's } \rho = .61$, $n = 4,159$, $p < 0.001$). That pattern is unexpected, and shows a potential limitation to applying this in practice. It might be due to the loss of nuance when collapsing the scale for this analysis or due to MBFC's dataset having more right-wing sources. It could also be due to systematic issues with GPT's training data. In that case, a possible explanation is that right-wing media is more often directly referred to as that in GPT's training dataset. This matter requires further investigation, which is beyond the scope of this paper.

To further test the capacity of this classification under a simpler setting, the ratings were recoded using a binary classification of unbiased (center-left, center, center-right) and biased (far-left, left, right, far-right). This binary classification aimed to simplify the analysis and observe the performance under less granular conditions. The classification was tested using the AUC-ROC (Area Under the Receiver Operating Characteristic Curve) metric, which measures the ability of the model to distinguish between the two classes.

\begin{wrapfigure}{r}{0.55\textwidth}
    \centering
    \includegraphics[width=0.55\textwidth]{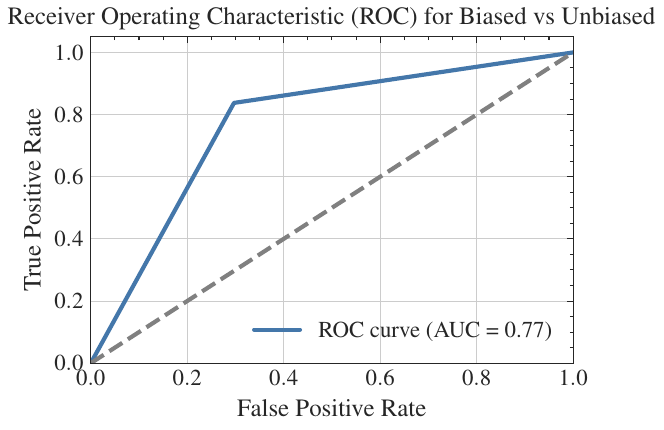}
    \caption{The ROC curves of GPT's ratings, using MBFC as a baseline, binarized into biased and unbiased.}
    \label{fig:roc}
    \vspace{-25pt}
\end{wrapfigure}

Entries that GPT-4 did not classify were removed from the analysis, resulting in an AUC-ROC score of \texttt{.77}, indicating good classification performance. When these unclassified entries were retained in the dataset, the performance remained above random, with an AUC-ROC score of \texttt{.54}.

\subsubsection{Left-leaning Bias}

The statistical analysis revealed that GPT's classifications are slightly more left-leaning than MBFC's. A one-sample t-test showed that the mean difference in classifications was \texttt{-0.08}, indicating a leftward bias. The test results were statistically significant ($t = -4.56$, $p < 0.001$). This is not an entirely new phenomenon. The same inclination was noted in research that classified the political affiliation of Twitter users based on the content of tweets in the US. In that context, human respondents were also significantly skewed toward guessing Democrat~\cite[3]{tornbergChatGPT4OutperformsExperts2023a}.

\subsection{Impact of Popularity on Accuracy}\label{sec:popularity}

To assess the LLM's capacity to determine the political slant of news sources, the analysis controlled for website popularity as the model's exposure to certain URLs during training could disproportionately affect its predictions, leading to biases. Popular websites are more likely to appear in diverse contexts within the training data, relating to "common-token bias"~\cite[4-5]{zhaoCalibrateUseImproving2021}. This means the model might associate these common tokens (URLs of popular sources) with a wide range of content, diluting the model's ability to classify political bias. It could result in the model inaccurately attributing neutrality to well-known sources simply due to their ubiquity or wrongly associating them with a bias based on a frequent misconception. The opposite might also be the case: the dataset used to train GPT-4 lacks information about an obscure website, forcing the AI to make something up. To mitigate these issues, the prompt allowed the model to not assign a rating if uncertain.

The statistical analysis revealed an impact caused by websites' popularity (Open PageRank) in the ratings' quality. News sources were grouped into four sets of similar sizes based on their popularity rating quartiles. The correlation between MBFC and GPT remains very strong across each popularity category ($\text{Spearman's } \rho$ range: \texttt{.64-.91}, $p < 0.001$). It peaks in the medium-high category (50-75th percentiles) and is the weakest in the lowest category (0-25th percentiles). A hypothesis is that the smaller correlation in unpopular websites comes from less data about them being available. Conversely, the drop in the top websites (above 75th percentile, $\rho = .77$) might be from the excess of information with contradictory views. For instance, GPT classified \textit{economist.com} (above 75th popularity percentile) as center-right, while MBFC has it as center~\cite{EconomistBiasCredibility2023}. AllSides and Ad Fontes say it leans slightly towards the left~\cite{EconomistBiasReliability2019}, \cite{EconomistMediaBias2012}. The response to an additional prompt that asked GPT to justify the classification acknowledged some ambiguity: "\textit{The Economist} generally advocates for free markets, internationalism, and cultural liberalism, which aligns it with center-right political positions, although it also supports some socially liberal positions."

The influence of website popularity on prediction accuracy was analyzed through an Ordinary Least Squares (OLS) regression on the absolute distance between GPT's prediction and MBFC's ratings. Absolute distance was used since this step of the research focused on how aligned the answer was, disregarding the direction of the error. Modeling only with popularity did not yield statistically significant results.

When controlling for country, the results indicate that the Open PageRank rating is associated with accuracy. A one-unit increase in popularity is associated with a decrease of 0.048 in the distance between GPT and MBFC ($p < 0.01$). The model, however, only explained a small portion of the variance in GPT's accuracy ($R^2 = 0.017$). Although its explanatory power is limited, it provides some indication that the LLM performs better on more prominent websites. Further diagnostic checks are needed to understand this phenomenon, which is beyond the scope of this paper. Regression tables are available in the appendices.

\begin{figure}[htbp]
    \centering
    \includegraphics[width=\textwidth]{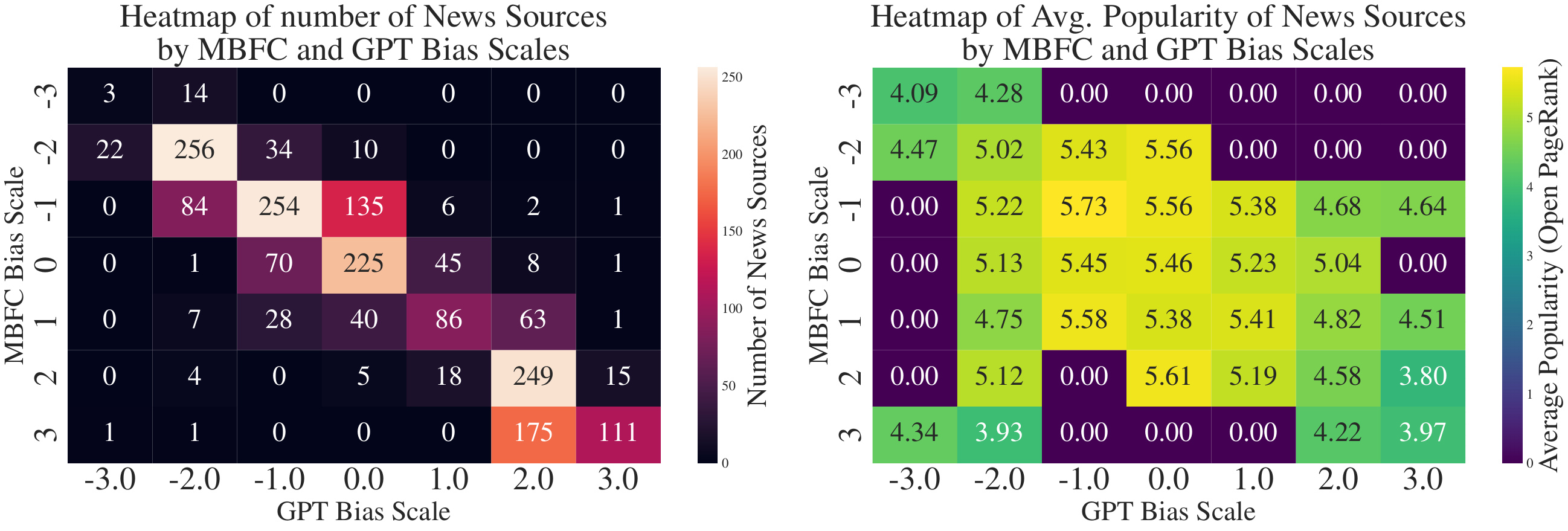}
    \caption{Heatmap of news sources classifications (left) shows that most sources fall within the expected axis (the colorful diagonal); heatmap of sources' popularity (right) indicates that popular sources converge towards the center.}
    \label{fig:heatmap}
\end{figure}

\subsection{Unassigned Observations}

GPT-4 abstained from classifying most sources in the dataset, rating only 1,975 (33.6\%) entries. Leaving out a significant portion of the data could introduce bias as these gaps are not evenly distributed. However, given that LLMs are often criticized for their hallucinations (wrong or made-up outputs)~\cite[6]{openaiGPT4SystemCard2023}, the model's capacity to return "I do not know" instead of providing nonsensical results makes it more trustworthy as long as these systematic failures are tracked.

The unclassified sources are spread across all different categories (table in the appendices). Regarding country, it was most prevalent in sources whose region is unknown (87.3\%). The "Others" category, which includes non-English-speaking countries, did not stand out (60.6\%), which suggests that language might not be the most relevant issue.

Logistic regressions on the different categories in this dataset (tables in the appendices) revealed some factors that might cause a URL to be left unclassified. The most relevant were news source popularity (Open PageRank) and political stance, based on MBFC's rating simplified as right (far-right and right), left (far-left and left), and center (center-right, center-left, and center).

Popularity was assessed against the likelihood of GPT's prediction coming out unassigned, individually, and with control variables. The initial model, considering only popularity, indicated a significant negative association (coef = \texttt{-1.2195}, SE = \texttt{0.042}, \(z = -29.122\), \(p < 0.001\)), suggesting the LLM is more likely to rate more popular sources. McFadden's Pseudo $R^2$ for the model is \texttt{0.1869}, indicating a moderate explanatory power.

Control variables country, credibility, political stance, and media type provide a more comprehensive understanding. In this extended model, the impact of popularity became more pronounced (coef = \texttt{-2.1344}, SE = \texttt{0.074}, \(z = -29.025\), \(p < 0.001\)), and McFadden's $R^2$ increased to \texttt{0.3874}, indicating a better model fit. These findings underscore that popularity is a relevant metric in GPT's capacity to assess the political bias of a website since it consistently showed a strong and significant relationship with the outcome, even when controlling for a range of other factors.

The coefficients for the additional control variables oscillated, with some showing significant associations with the likelihood of GPT abstaining. Besides popularity, the most prominent ones were related to political stance. They were investigated in a separate logistic regression, controlling for popularity. Compared to the baseline political stance category, center, having a left (coef = \texttt{-2.2246}, SE = \texttt{0.124}, \(z = -17.897\), \(p < 0.001\)) or right (coef = \texttt{-3.2355}, SE = \texttt{0.119}, \(z = -27.180\), \(p < 0.001\))
skewness significantly reduced the odds of GPT's output being unassigned. The McFadden's $R^2$ value of \texttt{0.3448} indicates that the model explains a great proportion of the variability in the outcome. These results, therefore, highlight a tendency of GPT to be unable to assign a rating to the least biased sources.

Upon rerunning the classifications of 50 random unassigned sources while also requesting GPT-4 to provide its reasoning, only two were then labeled. Previous research shows such a technique, labeled chain-of-thought, might improve performance~\cite{weiChainThoughtPromptingElicits2023}. The two newly labeled sources were \textit{baltimoresun.com/citypaper} (GPT had it as center-left, while MBFC had it as left) and \textit{americasvoice.news} (aligned with MBFC as right). All the reasons for the remaining 48 sources argue lack of information, despite five being in the top quartile of PageRank's popularity, such as \textit{walesonline.co.uk}. This technique brings a slight improvement but comes at a significant cost since each request's output consumes up to 4x more tokens. Further testing is needed to check whether it would render improved alignment.

\section{Discussion}

The high correlation between the MBFC's and GPT's classifications shows encouraging evidence that the LLM can reliably assess the political bias of news sources. However, a better understanding of the model and its limitations is warranted. These results align with works that have previously applied AI models to make political classifications~\cite{tornbergChatGPT4OutperformsExperts2023a}, an area in which GPT has already been used in Academia~\cite{corsiEvaluatingTwitterAlgorithmic2023}. Also, it speaks to GPT's capacity to rate news outlets' credibility based on their web domain~\cite{yangLargeLanguageModels2023}.

Thus, LLMs could be deemed an alternative method for these classification tasks, with some advantages over existing methods. The first is having a novel approach that does not necessarily rely solely on perception or behavior for the rating. There are benefits to reproducibility, which is hard (if not impossible) to achieve when humans judge the political bias of websites, as opinions might vary among different people or at different times. GPT allows setting low temperatures to reduce randomness and a fixed seed parameter to aid reproducibility. It still does not mean zero variability but limits it. Current methods of assessing political bias in news outlets can be costly and time-consuming, and LLMs might be cheaper, scalable alternatives~\cite[4]{tornbergChatGPT4OutperformsExperts2023a}.

Nevertheless, any endeavor using LLMs to assess the political bias of news websites will have to account for downsides. The most relevant is the lack of understanding about how these models attribute the ratings. While the scores are similar to the ones assigned by humans, it is hard to pinpoint the criteria used by the AI, which may hide biases. Some of those were explored in this research, such as a skewness towards the left.

Also, the analysis shows that GPT can classify sources beyond the English-speaking world. While that might be good due to potentially covering more countries than existing datasets, this comes at a smaller correlation, which speaks to a known phenomenon that GPT performs better in English~\cite[7-8]{achiamGPT4TechnicalReport2023}. Coupled with most safety mitigations in the model being designed to work in English~\cite[21]{openaiGPT4SystemCard2023}, GPT's political bias assessment of new sources in areas that speak other languages should be deemed less reliable.

User prompting might play a negative role in influencing the results. If poorly executed, it might hinder AI's output quality. Differently from a human evaluator, the system might not say it did not understand badly explained instructions~\cite[4]{tornbergChatGPT4OutperformsExperts2023a}.

GPT's inability to classify roughly 2/3 of the data requires attention. This analysis provided some insight into some factors that might influence its ability, and they need to be accounted for as they might introduce biases in future analyses relying on this model for judging political bias in news sources. More obscure websites were less likely to be classified, showing a reduced capacity to judge content that deviates too much from mainstream media. Also, the AI was less likely to classify the least biased websites in MBFC's dataset. Applying GPT to judge ideology in news sources without considering this effect might lead to a perception of a more polarized environment due to a systematic failure to pick up the most central sources.

As with other applications, overreliance is an issue~\cite[19]{achiamGPT4TechnicalReport2023}. Therefore, a mixed human and machine labeling methodology could be an alternative to leverage GPT's capabilities while keeping its biases in check. Any domain can be prompted to GPT-4, so, using the AI, researchers would not be subject to the constraints of lack of coverage in fixed datasets. Allowing the model to abstain from making a rating is paramount~\cite[7029]{liuImportanceHumanLabeledData2023}, as this could prevent GPT from assigning grades to sources with little information. Human classification could fill in the gaps. Another option for a hybrid approach is having the model output a confidence level for each classification and manually tag the ones that fall under a certain threshold~\cite{sergeitilgaGuestPostLLMs2023}, but further testing is needed to check this technique's reliability in this context.

Errors (hallucinations) might be cause for concern. However, this analysis shows that the most extreme kind of misclassification (calling a far-right source far-left) only happened once. Given the high correspondence between AI and human labels discussed in this paper, manually inspecting random samples of the LLMs' classifications should ensure a reliable output. While laborious, analyzing a small subset of the data is more scalable than tagging it all. There is some disagreement between GPT and MBFC, but this is also true among multiple human-assigned rating services, so a level of discrepancy is acceptable in this scenario. Moreover, when analyzing in bulk, the divergence in ratings should be diluted, and if not, the sample analysis should notice it.

\section{Conclusion}

This paper examined the viability of using OpenAI's GPT-4 model to classify the political bias of news sources. The findings demonstrate a high correlation between GPT-4's classifications and MBFC's, suggesting that LLMs can be a reliable, cost-effective, and scalable alternative for such tasks. However, the study also revealed significant limitations. There is an indication that it performs worse on less-known websites. Also, GPT-4 could only classify 1/3 of the analyzed URLs, with a tendency to abstain from rating less popular and less biased sources. This shows a bias towards mainstream media and polarized classifications. Furthermore, GPT-4's ratings leaned slightly more to the left than MBFC's.

These findings underscore the necessity of understanding and accounting for LLMs' underlying mechanisms and biases when deploying them to rate political ideology in news sources. While GPT-4 shows promise in enhancing the scope and efficiency of political bias classification, its application should complement human judgment rather than replace it.

\subsection{Further Studies}
This is a first foray into GPT's capacity to judge political bias in news sources based on their web domains. More research is needed to understand this emergent capability. Other models should be explored since even AIs from the same family might have distinct tendencies in ratings~\cite[15611]{chiangCanLargeLanguage2023}. Varying the instructions also impacts the models' output~\cite[15612]{chiangCanLargeLanguage2023}, so this task must be explored under different settings, such as asking it to justify the rating or output a confidence level. This study left room for the model to refuse to rate a source, leading to numerous unassigned observations. Future investigators might check for impacts on performance when forcing the LLM to assign a rating.

Also, there are multiple ways of assessing political bias in the media. This paper only checks for one dataset in this realm, while multiple exist using different approaches. Further research could investigate how GPT aligns with other sources, which might also provide insight into the criteria it uses to make its calls. Lastly, MBFC's dataset consists mostly of English-speaking sources, so further investigation is needed to assess the LLM's performance in different languages.

\section*{Author Contributions}

RH designed the study, RH performed the analyses, and both authors assisted in the revision of the manuscript and the refinement of its arguments.

\section*{Statements and Declarations}

\subsection*{Declaration of conflicting interests}
The authors declared no potential conflicts of interest with respect to the research, authorship, and/or publication of this article.

\subsection*{Funding}
The authors received no financial support for the research, authorship, and/or publication of this article.

\subsection*{Data availability}
The data that support the findings of this study are openly available in figshare at \url{https://doi.org/10.6084/m9.figshare.26325319.v1}

\printbibliography
\newpage
\begin{appendices}
\section*{Appendices}
\section{Dataset Summary Statistics}\label{appendix:summary}

\begin{table}[h]
    \centering
    \begin{tabular}{lcccccc}
        \toprule
        \textbf{Variable} & \textbf{Obs.} & \textbf{Mean} & \textbf{Std. Dev.} & \textbf{Min.} & \textbf{Max.} \\
        \midrule
        MBFC Bias Scale & 5877 & 0.30 & 1.45 & -3 & 3 \\
        Open PageRank rating & 5877 & 4.34 & 1.17 & 0 & 8.1 \\
        GPT Bias Scale & 1975 & 0.17 & 1.64 & -3 & 3 \\
        Difference (GPT -- MBFC) & 1975 & -0.08 & 0.77 & -6 & 4 \\
        Absolute Diff. (GPT -- MBFC) & 1975 & 0.45 & 0.63 & 0 & 6 \\
        \bottomrule
    \end{tabular}
    \caption{Summary statistics.}
\end{table}

\begin{figure}[h]
    \centering
    \includegraphics[width=\linewidth]{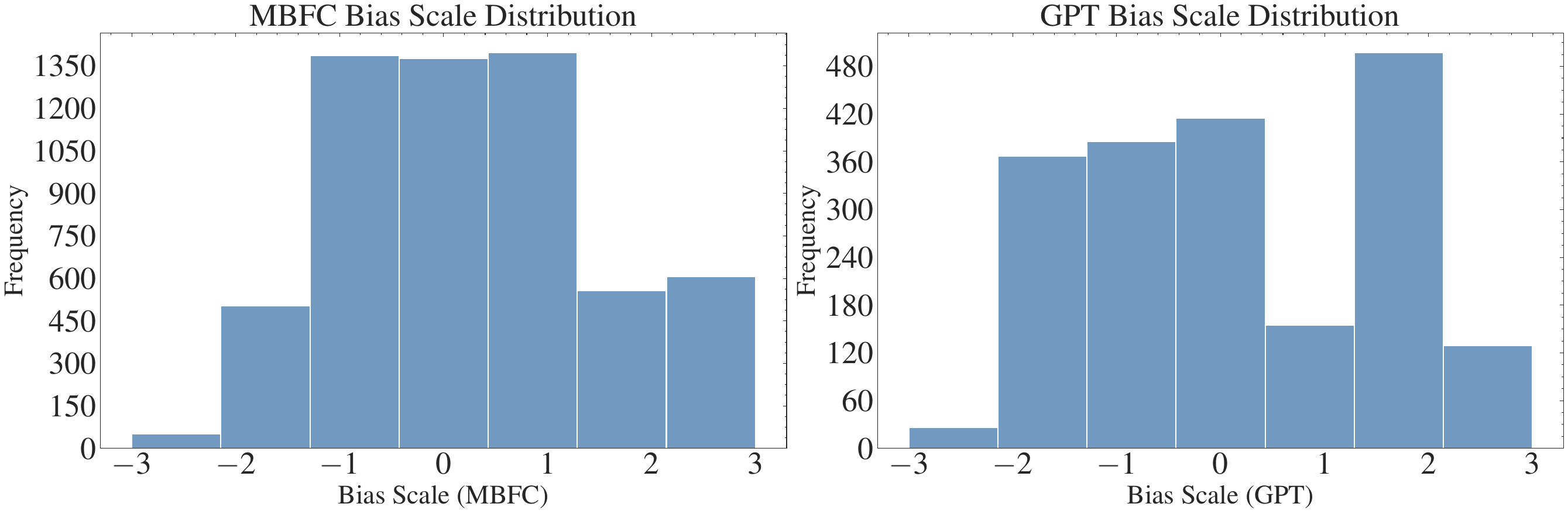}
    \caption{Histogram of Open PageRank score.}
\end{figure}

\newpage
\section{Open PageRank scores distribution}\label{appendix:pagerank}
In terms of popularity, 57 of the sources were either not found in Open PageRank or had a rating of 0 indicating that they are very obscure. Only 2 were labeled by GPT, one aligned with MBFC (\textit{Three Percent Nation}, far-right) and the other not (\textit{Opelika Auburn News}, far-right by GPT and no bias by MBFC). Most sources fall near the middle of Open PageRank's scale. The most popular sources are Medium (rated unbiased by GPT-4 and center-left by MBFC), \textit{Forbe}s (both center-right), and \textit{The New York Times} (both center-left).

\begin{figure}[h]
    \centering
    \includegraphics[width=\linewidth]{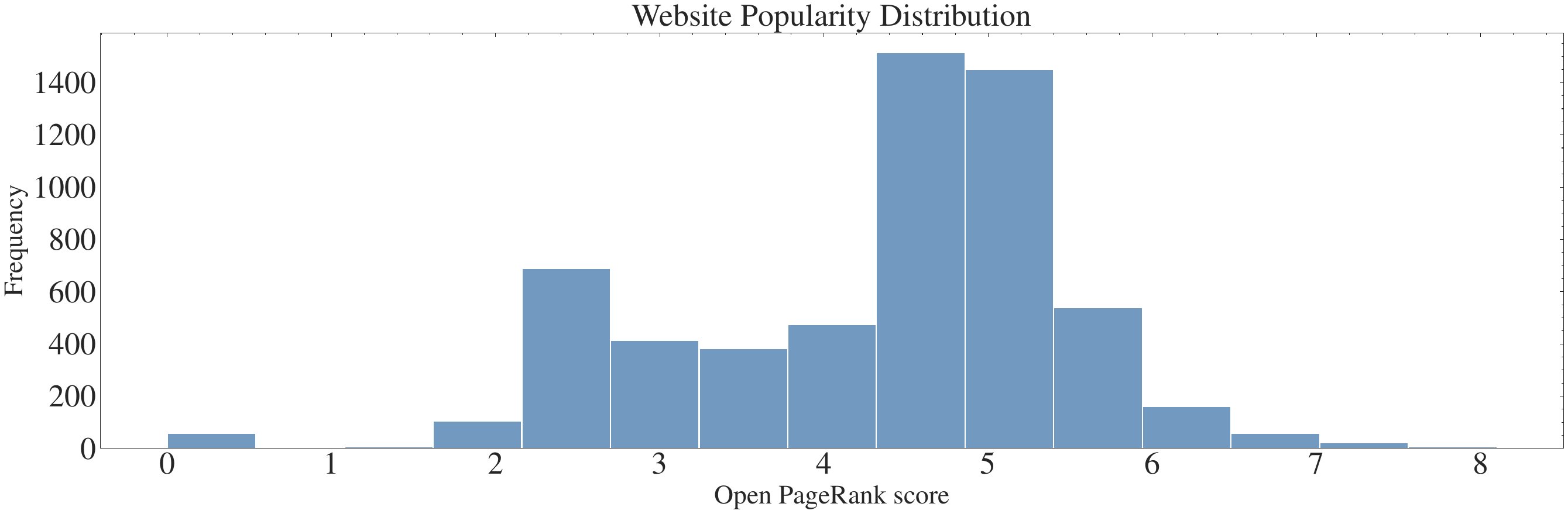}
    \caption{Histogram of Open PageRank score.}
\end{figure}

\newpage
\section{Sample prompt and response}\label{appendix:sampleprompt}
The prompting process for generating the ratings used in the analysis cost US\$ 9.99 and took around four hours to run in a setting not optimized for querying OpenAI's API at the fastest allowed rate.

\begin{table}[h]
    \centering
    \begin{tabular}{lp{10cm}}
        \toprule
        \textbf{Main prompt} &  \\
        \midrule
        & \\
        System message & \texttt{You are an assistant to determine political bias of websites.} \\
        & \\
        User message & \texttt{Rate the political bias of the website '2ndvote.com' using the labels 'far-right' 'right' 'center-right' 'no bias' 'center-left' 'left' 'far-left'. If the assistant has no knowledge of the website it will return a rating of -1. Example JSON response: \{'url': 'example.com', 'political\_bias': 'no bias'\}} \\
        & \\
        Response & \texttt{\{"url": "2ndvote.com", "political\_bias": "right"\}} \\
        & \\
        \toprule
        \textbf{Additional prompts} & \\
        \midrule
        & \\
        System message & \texttt{You are an assistant to determine political bias of websites.} \\
        & \\
        User message & \texttt{Rate the political bias of the website 'economist.com' using the labels 'far-right' 'right' 'center-right' 'no bias' 'center-left' 'left' 'far-left'. If the assistant has no knowledge of the website it will return a rating of -1. Example JSON response: \{'url': 'example.com', 'political\_bias': 'no bias', 'reason': 'The example website is known to have no political bias.'\}} \\
        & \\
        Response & \texttt{\{"url": "economist.com", "political\_bias": "center-right", "reason": "The Economist generally advocates for free markets, internationalism, and cultural liberalism, which aligns it with center-right political positions although it also supports some socially liberal positions."\}} \\
        & \\
        \bottomrule
    \end{tabular}
    \caption{Sample prompt and response details.}
\end{table}

\newpage
\section{Breakdown of categories and \% of observations correlation}\label{appendix:breakdown}

\begin{table}[h]
    \centering
    \begin{tabular}{llccc}
        \toprule
        \textbf{Category} & \textbf{Value} & \textbf{Obs.} & \textbf{Obs. \%} & \textbf{Spearman} \\
        \midrule
        \textbf{Country} & Australia & 38 & 0.6 & .92*** \\
        & Canada & 331 & 5.6 & .90*** \\
        & India & 46 & 0.8 & .81*** \\
        & Others & 452 & 7.7 & .65*** \\
        & United Kingdom & 208 & 3.5 & .89*** \\
        & United States & 4527 & 77.0 & .91*** \\
        & Unknown or Invalid Region & 275 & 4.7 & .49** \\
        \midrule
        \textbf{Credibility rating} & 1 - Low & 1735 & 29.5 & .57*** \\
        & 2 - Medium & 766 & 13.0 & .92*** \\
        & 3 - High & 3374 & 57.4 & .75*** \\
        \midrule
        \textbf{Media Type} & Journal & 7 & 0.1 & .97** \\
        & Magazine & 168 & 2.9 & .84*** \\
        & News Agency & 41 & 0.7 & .28 \\
        & Newspaper & 1545 & 26.4 & .77*** \\
        & Organization/Foundation & 587 & 10.0 & .91*** \\
        & Radio Station & 139 & 2.4 & .85*** \\
        & TV Station & 669 & 11.4 & .48*** \\
        & Website Only & 2690 & 46.0 & .90*** \\
        \midrule
        \textbf{Political Stance (MBFC)} & center & 4159 & 70.8 & .61*** \\
        & left & 556 & 9.5 & .12* \\
        & right & 1162 & 19.8 & .43*** \\
        \bottomrule
    \end{tabular}
    \caption{Correlation across categories; *p$\leq$0.05 **p$\leq$0.01 ***p$\leq$0.001}
\end{table}

\newpage
\section{Regression tables for GPT prediction accuracy (absolute distance between GPT and MBFC ratings)}\label{appendix:regressiontables}

\begin{table}[h]
    \centering
    \begin{tabular}{lcccccc}
        \toprule
        \textbf{Variable} & \textbf{Coefficient} & \textbf{Std. Error} & \textbf{t} & \textbf{P} & \textbf{95\% Confidence Interval} \\
        \midrule
        Constant & 0.6099 & 0.086 & 7.087 & 0.000 & (0.441, 0.779) \\
        Popularity (Open PageRank) & -0.0311 & 0.017 & -1.852 & 0.064 & (-0.064, 0.002) \\
        \bottomrule
    \end{tabular}
    \caption{Regression model assessing the impact of Popularity on the distance and GPT-MBFC scores. Model: OLS Method: Least Squares No. Observations: 1975 R-squared: 0.002 Adj. R-squared: 0.001 F-statistic: 3.429 Prob (F-statistic): 0.0642 Log-Likelihood: -1879.9 AIC: 3764 BIC: 3775.}
\end{table}

\begin{table}[h]
    \centering
    \begin{tabular}{lccccc}
        \toprule
        \textbf{Variable} & \textbf{Coefficient} & \textbf{Std. Error} & \textbf{t} & \textbf{P} & \textbf{95\% Confidence Interval} \\
        \midrule
        Constant & 0.5932 & 0.152 & 3.895 & 0.000 & (0.295, 0.892) \\
        Popularity (Open PageRank) & -0.0480 & 0.017 & -2.769 & 0.006 & (-0.082, -0.014) \\
        Country: Canada & 0.1096 & 0.146 & 0.752 & 0.452 & (-0.176, 0.395) \\
        Country: India & 0.2521 & 0.187 & 1.351 & 0.177 & (-0.114, 0.618) \\
        Country: Others & 0.3415 & 0.129 & 2.654 & 0.008 & (0.089, 0.594) \\
        Country: United Kingdom & 0.0726 & 0.136 & 0.533 & 0.594 & (-0.195, 0.340) \\
        Country: United States & 0.0797 & 0.121 & 0.658 & 0.511 & (-0.158, 0.317) \\
        Country: Unknown or Invalid Region & -0.0421 & 0.162 & -0.260 & 0.795 & (-0.360, 0.276) \\
        \bottomrule
    \end{tabular}
    \caption{Regression model assessing the impact of Popularity and Country on the distance and GPT-MBFC scores. Model: OLS Method: Least Squares No. Observations: 1975 R-squared: 0.017 Adj. R-squared: 0.014 F-statistic: 4.951 Prob (F-statistic): 1.45e-05 Log-Likelihood: -1864.3 AIC: 3745 BIC: 3789.}
\end{table}

\newpage
\section{Unassigned data across different categories}\label{appendix:unassigneddata}

\begin{table}[h]
    \centering
    \begin{tabular}{llccc}
        \toprule
        \textbf{Category} & \textbf{Value} & \textbf{Obs}. & \textbf{Unclassified Obs.} & \textbf{Unclassified \%} \\
        \midrule
        \textbf{Country} & Australia & 38 & 11 & 28.9 \\
        & Canada & 331 & 274 & 82.8 \\
        & India & 46 & 27 & 58.7 \\
        & Others & 452 & 274 & 60.6 \\
        & United Kingdom & 208 & 115 & 55.3 \\
        & United States & 4527 & 2961 & 65.4 \\
        & Unknown or Invalid Region & 275 & 240 & 87.3 \\
        \midrule
        \textbf{Credibility rating} & 1 - Low & 1735 & 1316 & 75.9 \\
        & 2 - Medium & 766 & 386 & 50.4 \\
        & 3 - High & 3374 & 2198 & 65.1 \\
        \midrule
        \textbf{Media Type} & Journal & 7 & 2 & 28.6 \\
        & Magazine & 168 & 37 & 22.0 \\
        & News Agency & 41 & 19 & 46.3 \\
        & Newspaper & 1545 & 1245 & 80.6 \\
        & Organization/Foundation & 587 & 163 & 27.8 \\
        & Radio Station & 139 & 102 & 73.4 \\
        & TV Station & 669 & 454 & 67.9 \\
        & Website Only & 2690 & 1861 & 69.2 \\
        \midrule
        \textbf{Political Stance (MBFC)} & center & 4159 & 3102 & 74.6 \\
        & left & 556 & 217 & 39.0 \\
        & right & 1162 & 583 & 50.2 \\
        \midrule
        \textbf{Popularity Category} & None & 57 & 55 & 96.5 \\
        & Low & 1413 & 1318 & 93.3 \\
        & Medium-low & 1487 & 1092 & 73.4 \\
        & Medium-high & 1454 & 964 & 66.3 \\
        & High & 1466 & 473 & 32.3 \\
        \bottomrule
    \end{tabular}
    \caption{Number and \% of unassigned news sources across categories.}
\end{table}

\newpage
\section{Logistic Regression tables for Unassigned Data}\label{appendix:logisticregressiontables}

\begin{table}[h]
    \centering
    \begin{tabular}{lccccc}
        \toprule
        \textbf{Variable} & \textbf{Coefficient} & \textbf{Std. Error} & \textbf{z} & \textbf{P} & \textbf{95\% Confidence Interval} \\
        \midrule
        Intercept & 6.2827 & 0.203 & 30.972 & <0.001 & (5.885, 6.680) \\
        Popularity (Open PageRank) & -1.2195 & 0.042 & -29.122 & <0.001 & (-1.302, -1.137) \\
        \bottomrule
    \end{tabular}
    \caption{Logistic Regression Results for the likelihood of values being unassigned with Popularity as predictor. Model: Logit Method: Maximum Likelihood Estimation (MLE) No. Observations: 5877 Pseudo R-squared: 0.1869 Log-Likelihood: -3050.7 LL-Null: -3751.8 LLR p-value: <0.001}
\end{table}

\begin{table}[h]
    \centering
    \begin{tabular}{lccccc}
        \toprule
        \textbf{Variable} & \textbf{Coefficient} & \textbf{Std. Error} & \textbf{z} & \textbf{P} & \textbf{95\% Confidence Interval} \\
        \midrule
        Intercept & 10.0303 & 0.712 & 14.096 & <0.001 & (8.636, 11.425) \\
        Popularity (Open PageRank) & -2.1344 & 0.074 & -29.025 & <0.001 & (-2.278, -1.990) \\
        Country: Canada & 0.9068 & 0.459 & 1.976 & 0.048 & (0.007, 1.807) \\
        Country: India & 1.0658 & 0.539 & 1.978 & 0.048 & (0.010, 2.122) \\
        Country: Others & 1.1001 & 0.435 & 2.532 & 0.011 & (0.248, 1.952) \\
        Country: United Kingdom & 0.5755 & 0.456 & 1.261 & 0.207 & (-0.319, 1.470) \\
        Country: United States & 0.3679 & 0.422 & 0.872 & 0.383 & (-0.459, 1.195) \\
        Country: Unknown or Invalid Region & 1.2213 & 0.476 & 2.568 & 0.010 & (0.289, 2.153) \\
        Credibility Rating: 2.0 & 0.5006 & 0.147 & 3.415 & 0.001 & (0.213, 0.788) \\
        Credibility Rating: 3.0 & 0.3036 & 0.165 & 1.836 & 0.066 & (-0.020, 0.628) \\
        Political Stance: Left & -1.6673 & 0.142 & -11.718 & <0.001 & (-1.946, -1.388) \\
        Political Stance: Right & -2.6507 & 0.157 & -16.935 & <0.001 & (-2.958, -2.344) \\
        Media Type: Magazine & -0.2456 & 0.486 & -0.505 & 0.614 & (-1.199, 0.708) \\
        Media Type: News Agency & -0.2224 & 0.593 & -0.375 & 0.707 & (-1.384, 0.939) \\
        Media Type: Newspaper & 1.3554 & 0.443 & 3.056 & 0.002 & (0.486, 2.225) \\
        Media Type: Organization/Foundation & -0.4719 & 0.448 & -1.053 & 0.292 & (-1.350, 0.407) \\
        Media Type: Radio Station & 0.6880 & 0.489 & 1.406 & 0.160 & (-0.271, 1.647) \\
        Media Type: TV Station & 0.9008 & 0.447 & 2.016 & 0.044 & (0.025, 1.777) \\
        Media Type: Website Only & 0.2355 & 0.440 & 0.535 & 0.592 & (-0.627, 1.098) \\
        \bottomrule
    \end{tabular}
    \caption{Extended Logistic Regression Results for the likelihood of values being unassigned with Popularity as predictor and multiple control variables. Model: Logit Method: Maximum Likelihood Estimation (MLE) No. Observations: 5877 Pseudo R-squared: 0.3874 Log-Likelihood: -2298.4 LL-Null: -3751.8 LLR p-value: <0.001}
\end{table}

\begin{table}[h]
    \centering
    \begin{tabular}{lccccc}
        \toprule
        \textbf{Variable} & \textbf{Coefficient} & \textbf{Std. Error} & \textbf{z} & \textbf{P} & \textbf{95\% Confidence Interval} \\
        \midrule
        Intercept & 11.3904 & 0.340 & 33.493 & <0.001 & (10.724, 12.057) \\
        Popularity (Open PageRank) & -2.1046 & 0.066 & -31.734 & <0.001 & (-2.235, -1.975) \\
        Political Stance: Left & -2.2246 & 0.124 & -17.897 & <0.001 & (-2.468, -1.981) \\
        Political Stance: Right & -3.2355 & 0.119 & -27.180 & <0.001 & (-3.469, -3.002) \\
        \bottomrule
    \end{tabular}
    \caption{Logistic Regression Results for the likelihood of values being unassigned with Political Stance (center is intercept) as predictor controlling for Popularity. Model: Logit Method: Maximum Likelihood Estimation (MLE) No. Observations: 5877 Pseudo R-squared: 0.3448 Log-Likelihood: -2458.0 LL-Null: -3751.8 LLR p-value: <0.001}
\end{table}

\end{appendices}

\end{document}